\documentclass{article}

\PassOptionsToPackage{numbers, compress}{natbib}

\usepackage{amsmath}
\usepackage{graphicx}
\usepackage{float}
\usepackage{makecell}
\usepackage[preprint]{neurips_2026}

\usepackage[utf8]{inputenc} 
\usepackage[T1]{fontenc}    
\usepackage{hyperref}       
\usepackage{url}            
\usepackage{booktabs}       
\usepackage{amsfonts}       
\usepackage{nicefrac}       
\usepackage{microtype}      
\usepackage{xcolor}         
\usepackage{multirow}

\usepackage{graphicx}
\usepackage{todonotes}
\usepackage[table]{xcolor}

\title{When Labels Have Structure: Improving Image Classification with Hierarchy-Aware Cross-Entropy}

\author{%
  April Chan$^{1}$ \\ \texttt{aprilc27@mit.edu}
  \And
  Davide D'Ascenzo$^{2,3}$ \\ \texttt{davide.dascenzo@unimi.it}
  \And
  Sebastiano Cultrera di Montesano$^{4}$ \\ \texttt{scultrer@broadinstitute.org}
}

\begin{document}

\maketitle

\renewcommand{\thefootnote}{}
\footnotetext{%
  \footnotesize
  $^{1}$Department of Electrical Engineering and Computer Science, Massachusetts Institute of Technology, Cambridge, MA, USA \quad
  $^{2}$Department of Computer Science, University of Milan, Milan, Italy \quad
  $^{3}$Department of Control and Computer Engineering, Politecnico di Torino, Torino, Italy \quad
  $^{4}$Eric and Wendy Schmidt Center, Broad Institute of MIT and Harvard, Cambridge, MA, USA
}
\renewcommand{\thefootnote}{\arabic{footnote}}

\begin{abstract}

Standard cross-entropy is the default classification loss across virtually all of machine learning, yet it treats all misclassifications equally, ignoring the semantic distances that a class hierarchy encodes. We propose Hierarchy-Aware Cross-Entropy (HACE), a drop-in replacement for standard cross-entropy that incorporates a known class hierarchy directly into the loss.
HACE combines two components: prediction aggregation, which propagates the model's probability mass upward through the class hierarchy to ensure that parent nodes accumulate the confidence of their children; and ancestral label smoothing, which distributes the ground-truth signal along the path from the true class to the root.
We evaluate HACE on CIFAR-100, FGVC Aircraft, and NABirds in two regimes: end-to-end training across six architectures spanning convolutional and attention-based designs, and linear probing on frozen DINOv2-Large features.
In end-to-end training, HACE improves accuracy over standard cross-entropy in 15 out of 18 architecture--dataset pairs, with a mean gain of 4.66\%.
In linear probing on frozen DINOv2-Large features, HACE outperforms all competing methods on all three datasets, with a mean improvement of 2.18\% over the next best baseline.
 
\end{abstract}

\section{Introduction}

Image classification is a foundational problem in machine learning, driving much of the field’s progress since its inception. Many of the most influential advances in modern machine learning have been developed in this context. In particular, convolutional neural networks (CNNs) were introduced to exploit the spatial structure of images~\cite{lecun1998gradient} and became the dominant paradigm following the success of AlexNet~\cite{krizhevsky2012imagenet}. Subsequent innovations, including deeper residual architectures~\cite{he2016deep}, normalization and regularization techniques~\cite{ioffe2015batch, srivastava2014dropout}, and large-scale data augmentation~\cite{shorten2019survey}, were largely motivated by improving image classification performance. More recently, attention-based architectures such as Vision Transformers~\cite{dosovitskiy2021an} and hierarchical transformers such as Swin~\cite{liu2021swin} have reached state-of-the-art results, alongside modernized convolutional designs such as ConvNeXt~\cite{liu2022convnet}.

Despite these advances, progress has been driven almost exclusively by improvements in architectures, optimization, and data. The classification objective itself has received far less attention, and standard cross-entropy remains the default choice in virtually every image classification pipeline. It treats all classes as independent, assigning equal penalty to every incorrect prediction, regardless of how semantically distant it is from the correct class. Yet this is rarely how we think about errors. A model that confuses a husky with a beagle has made a smaller mistake than one that confuses it with a Persian cat, since the first two are both dogs while the third belongs to a different species. Confusing a husky with a motorcycle is a bigger mistake still. The hierarchy that makes these distinctions meaningful is often known in advance, encoded in dataset taxonomies or biological ontologies, yet standard cross-entropy discards it entirely.


In this work, we revisit the classification objective and incorporate hierarchy directly into the loss. We propose \textbf{Hierarchy-Aware Cross-Entropy (HACE)} as a drop-in replacement for standard cross-entropy whenever a hierarchy is available. HACE combines two components: \emph{prediction aggregation}, which propagates model probability mass upward so that the score at each node reflects confidence in the entire subtree below it; and \emph{ancestral label smoothing}, which constructs a soft ground-truth distribution by placing decreasing probability mass at each ancestor of the true leaf class. We show in Section~\ref{sec:methods} that neither component yields a proper hierarchical loss on its own when labels are available only at the leaf
level---which is the standard setting in image classification---and that using them together resolves this.

The prediction aggregation mechanism was introduced by Cultrera di Montesano et al.~\cite{cultrera2026} for cell-type annotation, where training labels exist at multiple levels of a biological ontology. We adapt it to the leaf-only supervision regime and pair it with  ancestral label smoothing to provide a training signal at every level of the hierarchy.

We conduct experiments on three datasets with hierarchical label structures---CIFAR-100~\cite{krizhevsky2009cifar},
FGVC Aircraft~\cite{maji2013fgvc}, and NABirds~\cite{vanhorn2015nabirds}---in two regimes: end-to-end training across six architectures spanning convolutional and attention-based designs, and linear probing on top of frozen DINOv2-Large~\cite{oquab2024dinov2} features.
In end-to-end training, HACE improves accuracy over standard cross-entropy in 15 out of 18 architecture--dataset pairs, with a mean gain of 4.66\%. In linear probing, HACE outperforms all competing methods on all three datasets, with a mean improvement of 2.18\% over the next best baseline.
A comparative analysis shows that applying standard label smoothing prior to ancestral label smoothing further improves performance in most configurations, and that both variants outperform competing baselines including HXE and structure-aware soft labelling~\cite{bertinetto2020making}, which we describe in Section~\ref{sec:related}. Our results suggest that incorporating hierarchy at the level of the loss is a complementary direction to architectural improvement.


Our main contributions are threefold. First, we propose HACE, a loss that encodes the class hierarchy into both the predicted and target distributions, and establish why the two components must be used together in the leaf-only supervision setting. Second, we provide the first large-scale evaluation of a combined aggregation-and-smoothing hierarchical loss for image classification, covering six architectures, three datasets, and two training regimes. Finally, a comparative analysis demonstrates that standard label smoothing applied prior to ancestral smoothing brings further gains, and that HACE outperforms standard cross-entropy, HXE, and standard label smoothing on the large majority of configurations.

\section{Related Work}
\label{sec:related}

\paragraph{Prediction aggregation and hierarchical losses.}
Prediction aggregation for structured label spaces was introduced by Cultrera di Montesano et al.~\cite{cultrera2026} as the Hierarchical Cross-Entropy (HCE) loss for cell-type annotation in single-cell RNA sequencing. In atlas-scale annotation, the training data contain labels at multiple levels of the cell ontology simultaneously: some cells are annotated as \emph{T cell} while others are labelled at the finer level \emph{CD4$^+$ T cell}.
Standard cross-entropy treats these as independent classes and does not enforce that the probability assigned to a parent is at least the sum of its children.
HCE resolves this by aggregating predicted probabilities up the ontology, so that predicting a fine-grained subtype implies endorsing all its ancestors.
Because the training data carry ground-truth labels at every annotated level, the loss is well-defined without any modification of the targets, and HCE was shown to improve performance across architectures.

This is not the case in image classification, where ground-truth labels are usually assigned at the leaf level only. There are no annotations at internal nodes, so applying prediction aggregation in isolation provides model outputs at internal nodes but no training signal for them. HACE closes this gap by pairing aggregation with ancestral label smoothing.

A related loss is HXE~\cite{bertinetto2020making}, proposed by Bertinetto et al. for classification tasks where classes are organized into a known taxonomy.
HXE modifies the loss by computing, from a standard global softmax over leaf classes, the implied conditional probabilities at each level of the tree via their algebraic relationship to the leaf probabilities, and penalising errors at each level with an exponentially decaying weight toward the root. This enriches the loss with hierarchical structure, but does not modify the target distribution; the ground truth remains a one-hot vector at the leaf. HACE differs in that it modifies both sides of the cross-entropy, providing a supervision signal at every node in the tree, not just the leaf.
We include HXE as a baseline in our study.

Earlier work on hierarchical classification often relied on tree-structured classifiers~\cite{silla2011survey} or cost-sensitive losses that penalize predictions in proportion to their distance in the taxonomy~\cite{deng2010classifying}. These approaches either require separate classifiers at each level or depend on a hand-crafted distance function, whereas HACE is a single differentiable loss that requires only the tree structure and no additional design choices.

\paragraph{Label smoothing and soft targets.}
Label smoothing~\cite{szegedy2016rethinking} replaces the one-hot encoding with a mixture of the true class and a uniform distribution.
It improves calibration and reduces overconfidence, but spreads mass uniformly across all classes with no regard for their relationships.
M\"{u}ller et al.~\cite{muller2019does} study when label smoothing helps and show that it acts as a regularizer, but also find that it can hurt knowledge distillation by erasing structure from the penultimate layer's representations.
A more structured alternative is to distribute the smoothing mass according to class similarity.
Bertinetto et al.~\cite{bertinetto2020making} propose soft labels in which the probability assigned to each class decays exponentially with the height of the lowest common ancestor (LCA) between that class and the ground truth, producing targets that are more informative about class relationships than uniform smoothing.
Kobs et al.~\cite{kobs2020simloss} propose SimLoss, which constructs soft targets by distributing probability mass across all classes according to a pairwise similarity matrix derived from label co-occurrence or semantic embeddings, generalizing structured smoothing to settings where no explicit taxonomy is available.
Peterson et al.~\cite{peterson2019human} derive soft labels from human annotator disagreements on CIFAR-10, showing that targets reflecting genuine labelling uncertainty improve robustness.
Our ancestral label smoothing belongs to this family but differs in two respects: mass is placed only along the ancestral path of the true class rather than across all classes, and the decay is geometric with depth rather than LCA-based.
Crucially, as we discuss in Section~\ref{sec:methods}, ancestral label smoothing only yields a proper hierarchical loss when paired with prediction aggregation, since the targets at internal nodes require corresponding model outputs to compare against.

\paragraph{Hierarchical representation learning.}
A separate line of work incorporates hierarchy through the representation rather than the loss.
Barz and Denzler~\cite{barz2019hierarchy} derive a class embedding from the hierarchy such that cosine distances between embeddings reflect semantic similarity, and train models to reproduce those embeddings. Khrulkov et al.~\cite{khrulkov2020hyperbolic} embed images in hyperbolic space, where the geometry naturally reflects tree-structured relationships. Both approaches require either a pre-computed class embedding or a change to the metric space of the output layer; HACE works with any existing architecture without modification, provided that a known class hierarchy is available.

\begin{figure}
    \centering
    \includegraphics[width=0.65\linewidth]{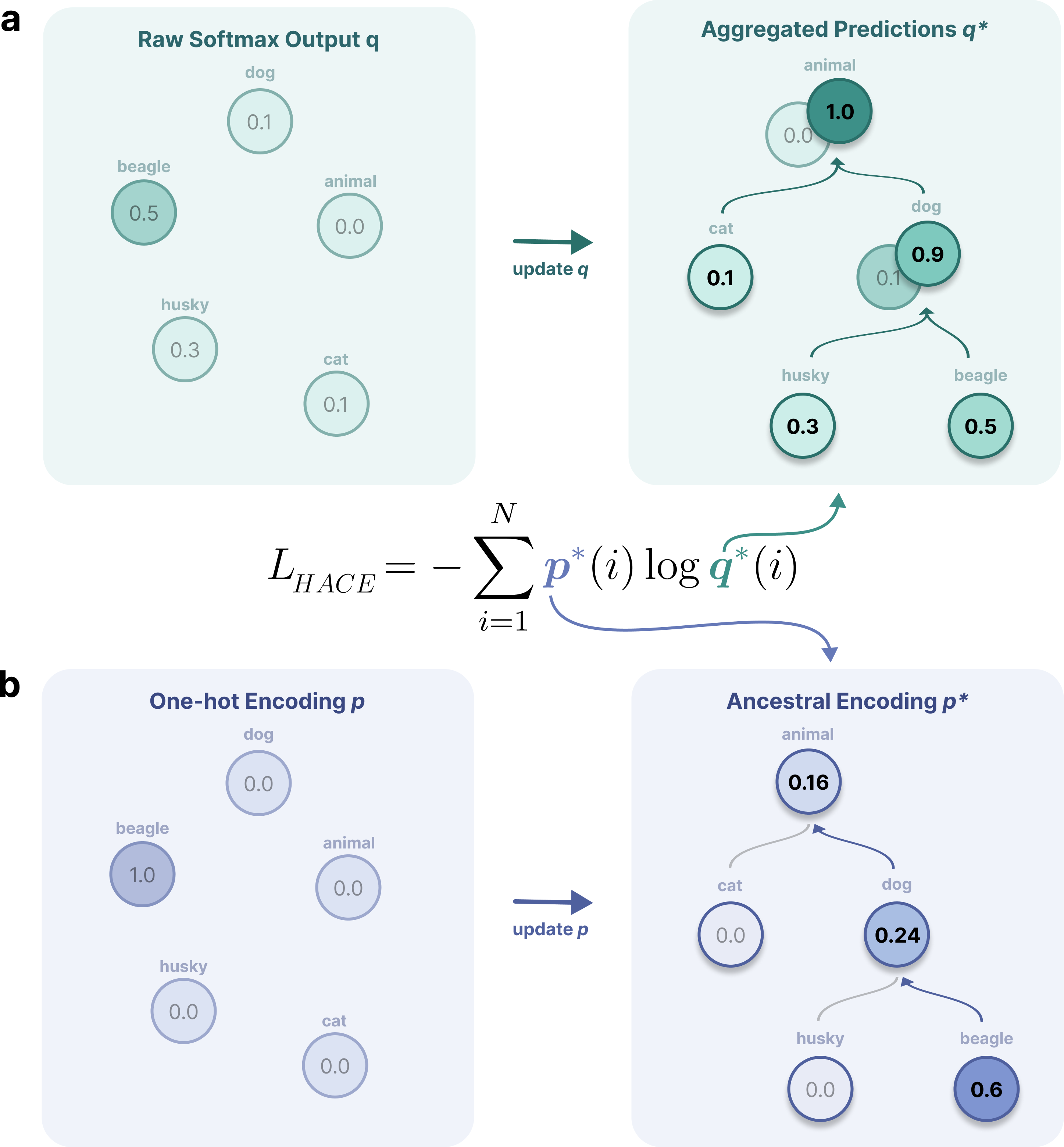}
    \caption{Illustration of the two components of HACE applied to a toy animal hierarchy. \textbf{(a)} \textbf{Prediction aggregation (updating $q$).} The raw softmax output $q$ assigns probabilities to each class independently. Aggregated predictions $q^*$ are obtained by propagating mass upward through the tree according to Equation~(\ref{eq:aggregation}), so that each internal node accumulates the confidence of its entire subtree. \textbf{(b)} \textbf{Ancestral label smoothing (updating $p$).} The one-hot encoding $p$ assigns all probability mass to the ground-truth class and zero mass to all other nodes. The ancestral encoding $p^*$ is constructed by propagating mass upward along the path from true class to the root using dilution $d = 0.6$.}
    \label{fig:PandQ}
\end{figure}

\section{Methods}
\label{sec:methods}

Both standard cross-entropy (SCE) and HACE minimize the cross-entropy between a ground-truth distribution $p$ and a predicted distribution $q$:
\begin{equation}
  \mathcal{L} = -\sum_{i} p(i) \log q(i).
  \label{eq:xent}
\end{equation}
They differ only in how $p$ and $q$ are constructed.
SCE uses a one-hot $p$ over $n$ leaf classes and a flat softmax $q$ over the same $n$ classes.
HACE expands both to cover all $N$ nodes in the class hierarchy.

We require a class hierarchy organized as a tree\footnote{Technically, the method extends straightforwardly to directed acyclic graphs (DAGs), since both prediction aggregation and ancestral label smoothing only depend on the ancestor--descendant relationships between nodes. All hierarchies considered in this paper are trees, so we use tree terminology throughout for simplicity. See Appendix~\ref{subsec:dag_extension} for the complete discussion.}, with $n$ leaf nodes and $N$ total nodes.
In practice, each dataset comes with a taxonomy; we extract it as a directed tree with edges pointing from parent to child.

Modifying the loss to incorporate hierarchy requires updating both sides of Equation~(\ref{eq:xent}).
We update $q$ through \emph{prediction aggregation} (Section~\ref{subsec:aggregated_prediction}), which propagates probability mass \emph{upward} through the tree so that every internal node accumulates the confidence of its subtree. We update $p$ through \emph{ancestral label smoothing}
(Section~\ref{subsec:label_smoothing}), which redistributes the ground-truth signal \emph{along the ancestral path} of the true class.
These two modifications play complementary roles, and neither is sufficient on its own in the leaf-only supervision regime; we discuss why at the end of each subsection.

\subsection{Prediction Aggregation}
\label{subsec:aggregated_prediction}

The model output layer is expanded to $N$ units, one per node in the class hierarchy.
Given raw logits $z \in \mathbb{R}^N$, we compute a softmax $\hat{q} = \mathrm{softmax}(z) \in \mathbb{R}^N$. Probability mass is then aggregated upward:
\begin{equation}
  q^*(i) = \hat{q}(i) + \sum_{j \in \mathcal{D}(i)} \hat{q}(j),
  \label{eq:aggregation}
\end{equation}
where $\mathcal{D}(i)$ is the set of all descendants of node $i$.
Thus $q^*(i)$ is the model's total probability that the true class is $i$ or any of its subtypes.
The aggregation is applied to every node $i \in \{1, \ldots, N\}$, yielding $q^* \in \mathbb{R}^N$. See Figure~\ref{fig:PandQ}a for an example and Appendix~\ref{subsec: Reachability Matrix} for computational details.


\paragraph{Why aggregation alone is not enough.}
With leaf-only supervision, the one-hot target $p$ is zero at every internal node.
The cross-entropy loss at those positions is zero regardless of $q^*$, so the aggregated outputs at internal nodes receive no gradient.
The hierarchy is invisible to the loss, an issue we address in the next section by providing a training signal at every level of the hierarchy.

\subsection{Ancestral Label Smoothing}
\label{subsec:label_smoothing}

Existing soft-target methods modify $p$ by redistributing probability mass \emph{across the leaf classes}.
Standard label smoothing~\cite{szegedy2016rethinking} spreads mass uniformly over all $n$ leaves; the structured soft labels of Bertinetto et al.~\cite{bertinetto2020making} concentrate it on leaves that are semantically close to the true class, using LCA-distance as the measure of closeness.
Both operate \emph{horizontally}: the target remains a distribution over leaf classes, and no supervision signal is introduced at internal nodes of the hierarchy.

Ancestral label smoothing takes a different axis.
Rather than redistributing mass among leaves, it propagates mass \emph{vertically}, from the true leaf upward along its ancestral path to the root.
This turns the scalar ground-truth signal into a supervision signal that is present at \emph{every level} of the hierarchy, not only at the leaves.
The key consequence is that meaningful cross-entropy terms appear at internal nodes, which is why ancestral label smoothing and prediction aggregation must be used together: one produces targets at internal nodes, the other produces model outputs at those same nodes.

\emph{Horizontal} and \emph{vertical} smoothing are complementary, and the two-step construction below is designed to exploit both.
Step~1 applies an optional horizontal redistribution over the leaves; Step~2
then propagates whatever mass has been assigned to each leaf upward along its ancestral path.

Let $a_0$ be the ground-truth leaf class.
We construct a soft target $p^* \in \mathbb{R}^N$ in two steps.

\paragraph{Step 1: standard label smoothing (optional, horizontal).}
We first apply standard label smoothing~\cite{szegedy2016rethinking} to the one-hot target, mixing it with a uniform distribution over all $n$ leaf classes using a smoothing parameter $\varepsilon \in [0, 1)$:
\begin{equation}
\tilde{p}(i) = (1 - \varepsilon)\,\mathbf{1}[i = a_0] + \frac{\varepsilon}{n},
\quad i \in \{1, \ldots, n\}.
\label{eq:uniform_smooth}
\end{equation}
This step is optional; setting $\varepsilon = 0$ recovers a one-hot and restricts all smoothing to the vertical axis.
The uniform distribution in Equation~\ref{eq:uniform_smooth} can be also replaced by the LCA-based soft labels of Bertinetto et al.~\cite{bertinetto2020making}, which concentrate the horizontal mass on leaves that are semantically close to the true class. Both horizontal smoothings are evaluated in our comparative analysis (Section~\ref{subsec:ablation}).

\paragraph{Step 2: ancestral label smoothing (vertical).}
We introduce a dilution parameter $d \in (0,1]$ and apply ancestral redistribution to the entire distribution $\tilde{p}$.
For each leaf $c$ with mass $\tilde{p}(c) > 0$, let $b_0, b_1, \ldots, b_k$ denote the path from $c$ to the root, where $b_0 = c$.
The contribution of leaf $c$ to node $b_j$ on its ancestral path is:
\begin{equation}
  p^*(b_j) \mathrel{+}= d\,(1-d)^j\,\tilde{p}(c) \quad \text{for } j < k,
  \qquad
  p^*(b_k) \mathrel{+}= \tilde{p}(c) - \sum_{j=0}^{k-1} d\,(1-d)^j\,\tilde{p}(c),
  \label{eq:smoothing}
\end{equation}
where contributions from all leaves are summed at shared ancestors.
All non-ancestral internal nodes receive zero mass.
The mass at each leaf decays geometrically as it propagates upward, so most weight stays near the leaves and progressively less reaches coarser ancestors.
When $d = 1$ no mass is propagated and $p^*$ reduces to $\tilde{p}$.
When $\varepsilon = 0$, only the ground-truth leaf $a_0$ has nonzero mass in $\tilde{p}$, and the construction reduces to a purely ancestral distribution along the path of $a_0$.
Figure~\ref{fig:PandQ}b illustrates $p^*$ for a toy hierarchy with $d = 0.6$ and $\varepsilon = 0$.

\paragraph{Why ancestral smoothing alone is not enough.}
With ancestral label smoothing, the soft target $p^*$ places mass on both a leaf class and its ancestors, but without prediction aggregation the model has no mechanism to ensure consistency between them: it is implicitly asked to treat a class and its own superset as competing alternatives.
This is the same inconsistency identified by Cultrera di Montesano et al.~\cite{cultrera2026}, where assigning independent probabilities to \emph{T cell} and \emph{CD4$^+$ T cell} violates the basic constraint that the probability of a parent must be at least as large as the probability of any of its children.
Prediction aggregation enforces this constraint by construction, since $q^*(i) \geq q^*(j)$ for any descendant $j$ of $i$.

\section{Experimental Setup}

In this section, we introduce the models, datasets and training procedure considered in this paper.

\subsection{Models}

To ensure a thorough and fair evaluation of HACE, we select six architectures spanning two design families.
On the convolutional side, we include the ResNet family at three depths---ResNet-18, ResNet-34, and ResNet-50~\cite{he2016deep}---to examine how model capacity interacts with the hierarchical loss, alongside ConvNeXt~\cite{liu2022convnet}, a modernized convolutional design that matches attention-based models on standard benchmarks.
On the attention-based side, we include ViT-B/16~\cite{dosovitskiy2021an}, the first pure transformer architecture to outperform convolutional models at scale, and the Swin Transformer~\cite{liu2021swin}, which introduces a hierarchical window-attention scheme that bridges the two design families.
Together, these six architectures provide a representative cross-section of the current model landscape and allow us to assess whether the benefits of HACE are consistent across different inductive biases.

In addition to these six architectures trained end-to-end, we evaluate HACE in a linear probing setting using DINOv2-Large~\cite{oquab2024dinov2} as a fixed feature extractor.
DINOv2-Large is a ViT-L/14 trained with self-supervised DINO objectives on a large curated dataset; we use it as a frozen backbone and train only a linear classifier on top of its features, as described in Section~\ref{subsec:finetuning}.

For a complete description of the models, see Appendix~\ref{app:models}.

\subsection{Datasets}

We evaluate HACE on three datasets with hierarchical label structures that differ in size, depth, and regularity of their taxonomies, as summarized in Table~\ref{tab:datasets}. A detailed description of each dataset, including the structure of its hierarchy and any preprocessing steps, is provided in Appendix~\ref{app:datasets}.

\begin{table}[H]
\caption{Summary of the three evaluation datasets. Hierarchical depth counts
levels below the root, excluding the root itself.}
\label{tab:datasets}
\renewcommand\theadfont{\normalsize}
\centering
\begin{tabular}{l c c c c c c}
\toprule
\textbf{Dataset} & \textbf{\thead{ Train Set \\ Size}} & \textbf{\thead{Test Set \\ Size}} & \textbf{\thead{Leaf \\ Classes}} & \textbf{\thead{Total \\ Classes}} & \textbf{\thead{Hierarchical \\ Depth}} & \textbf{\thead{Cropped Image \\ Size}} \\
\midrule
\rowcolor{gray!10} CIFAR-100 & 50,000 & 10,000 & 100 & 120 & 2 & $32\times32$ \\
FGVC & 6,800 & 3,400 & 102 & 201 & 3 & $224\times224$ \\
\rowcolor{gray!10} NABirds & 23,929 & 24,633 & 555 & 1011 & 3 or 4 & $224\times224$ \\
\bottomrule
\end{tabular}
\end{table}

\subsection{Training procedure}

We evaluate HACE in two regimes.
In \emph{end-to-end training}, all weights are optimized from random initialization on each dataset.
In \emph{linear probing}, all images are first passed through the frozen DINOv2-Large backbone; a single linear classifier is then trained on the resulting fixed features.

For end-to-end training, we run every combination of six architectures, three datasets, and three dilution values ($d \in \{0.2, 0.5, 0.7\}$) under both HACE and SCE.
Each architecture--dataset combination is trained with six HACE models, corresponding to three dilution values each under two learning rate pairings, and four SCE models: one at the standard learning rate and one adjusted to
match the leaf-level gradient magnitude of each dilution value; full details are given in Appendix~\ref{app:lr}.
HACE and SCE follow identical forward passes, with one exception: the HACE output layer produces $N$ logits over all hierarchy nodes rather than $n$ leaf logits.

All hyperparameters are set to the values recommended by the original authors and held fixed throughout. Any observed differences in performance can therefore be attributed to the choice of loss function alone, and the reported results represent a empirical lower bound on what HACE could achieve with dedicated hyperparameter tuning.

\subsection{Evaluation metrics}

We report two metrics for all experiments.
\textbf{Top-1 accuracy} is the fraction of test samples for which the highest-scoring leaf matches the ground truth; \textbf{Top-5 accuracy} is the fraction for which the ground truth appears among the five highest-scoring leaves.
Both metrics are computed exclusively over leaf-level predictions, ensuring comparability between HACE---whose output layer spans all $N$ nodes---and SCE, which produces logits over the $n$ leaf classes only. Top-1 accuracy results are reported in the main body; top-5 results are provided in Appendix~\ref{app:top5}.

In Section~\ref{subsubsec:hierarchical} we additionally examine per-class
accuracy at the family level of the FGVC hierarchy, by aggregating leaf
probabilities upward and measuring how often the correct family node is
selected.
We report the corresponding analysis at the manufacturer level in
Appendix~\ref{app:hierarchical_fgvc}.
For SCE, which produces no internal-node outputs, this aggregation is applied post-hoc at evaluation time; HACE already produces aggregated outputs by construction.

\section{Results}
\label{sec:results}

We compare the performance of HACE and SCE across a wide range of architectures, datasets, and dilution values in both end-to-end training and linear probing settings.

\subsection{End-to-end training}

Table~\ref{tab:besttobest} reports, for each architecture--dataset pair, the best top-1 accuracy achieved by HACE and SCE across all dilution values. HACE outperforms SCE in 15 out of 18 pairs, with a mean gain of 4.66\% and a maximum decrease of 1.57\%.
The gains are consistent across both convolutional and attention-based designs and grow with the complexity of the hierarchy: improvements are largest on NABirds and FGVC, where the taxonomy is deeper and more irregular than on CIFAR-100.

\begin{table}[H]
    \centering
    \caption{Top-1 accuracy (\%) of HACE and SCE on end-to-end training, for each architecture--dataset pair. Each cell reports the best result across all dilution values $d \in \{0.2, 0.5, 0.7\}$ and both learning rate pairings. The better result per pair is \textbf{bolded}.}
    \label{tab:besttobest}
    \begin{tabular}{l cc cc cc}
        \toprule
        & \multicolumn{2}{c}{\textbf{CIFAR-100}}
        & \multicolumn{2}{c}{\textbf{FGVC}}
        & \multicolumn{2}{c}{\textbf{NABirds}} \\
        \cmidrule(lr){2-3} \cmidrule(lr){4-5} \cmidrule(lr){6-7}
         & HACE & SCE & HACE & SCE & HACE & SCE \\
        \midrule
        \rowcolor{gray!10} ResNet-18   & \textbf{77.93} & 76.52 & \textbf{78.88} & 68.29 & \textbf{67.93} & 62.94 \\
        ResNet-34   & 78.01 & \textbf{78.30} & 77.32 & \textbf{77.74} & 69.44 & \textbf{71.01} \\
        \rowcolor{gray!10} ResNet-50   & \textbf{80.53} & 79.86 & \textbf{80.62} & 69.88 & \textbf{67.90} & 59.07 \\
        ConvNeXt-T    & \textbf{44.48} & 43.30 & \textbf{38.82} & 30.45 & \textbf{50.74} & 30.35 \\
        \rowcolor{gray!10} ViT-B/16    & \textbf{64.45} & 63.56 & \textbf{20.67} & 15.72 & \textbf{29.16} & 27.44 \\
        Swin-T      & \textbf{59.44} & 56.72 & \textbf{44.34} & 38.16 & \textbf{44.01} & 41.51 \\
        \bottomrule
    \end{tabular}
\end{table}


\subsubsection{Model-matched comparisons}

Table~\ref{tab:model_matched} provides a more granular analysis on the FGVC dataset, reporting top-1 accuracy for every combination of architecture and dilution value under both standard and adjusted learning rate pairings. HACE outperforms SCE in 31 out of 36 configurations, with an average improvement of 8.83\%. Three SCE runs---ResNet-18 at $d=0.7$, ResNet-34 at $d=0.5$, and ResNet-34 at $d=0.7$---produced near-zero accuracies. As detailed in Appendix~\ref{app:lr}, these correspond to pairings at high dilution values, where the SCE learning rate is scaled down to match the reduced leaf-level gradient magnitude of HACE, making training instability possible.
Excluding these runs, HACE outperforms SCE by a mean of 3.31\%, with a maximum gain of 21\%.
\begin{table}[H]
    \centering
    \caption{Top-1 accuracy (\%) of HACE and SCE on FGVC Aircraft, broken down by architecture, dilution value $d$, and learning rate pairing (standard / adjusted). Each HACE model is compared against the SCE model whose learning rate has been scaled to match the leaf-level gradient magnitude of that dilution value (see Appendix~\ref{app:lr}). The better result per configuration is \textbf{bolded}.}
    \label{tab:model_matched}
    \resizebox{\textwidth}{!}{%
    \begin{tabular}{l *{6}{cc}}
        \toprule
        & \multicolumn{2}{c}{$d = 0.2$}
        & \multicolumn{2}{c}{$d = 0.2$ adj}
        & \multicolumn{2}{c}{$d = 0.5$}
        & \multicolumn{2}{c}{$d = 0.5$ adj}
        & \multicolumn{2}{c}{$d = 0.7$}
        & \multicolumn{2}{c}{$d = 0.7$ adj} \\
        \cmidrule(lr){2-3} \cmidrule(lr){4-5} \cmidrule(lr){6-7}
        \cmidrule(lr){8-9} \cmidrule(lr){10-11} \cmidrule(lr){12-13}
        
            & HACE & SCE & HACE & SCE & HACE & SCE
            & HACE & SCE & HACE & SCE & HACE & SCE \\
        \midrule
        \rowcolor{gray!10} ResNet-18 & \textbf{74.11} & 61.36 & \textbf{78.94} & 69.37 & \textbf{71.23} & 67.45 & \textbf{71.59} & 69.37 & \textbf{70.18} & 3.39 & \textbf{71.44} & 69.37 \\
        ResNet-34 & 72.73 & \textbf{78.34} & \textbf{77.83} & 71.98 & \textbf{75.97} & 2.13 & \textbf{73.60} & 71.98 & \textbf{72.94} & 5.01 & \textbf{73.30} & 71.98 \\
        \rowcolor{gray!10} ResNet-50 & \textbf{72.28} & 61.90 & \textbf{81.10} & 71.44 & \textbf{69.79} & 69.13 & \textbf{76.78} & 71.44 & \textbf{72.76} & 70.45 & 70.78 & \textbf{71.44} \\
41.36 - 36.33
ConvNeXt-T & \textbf{34.14} & 13.14 & \textbf{32.31} & 31.23 & \textbf{31.89} & 29.10 & \textbf{39.42} & 31.23 & \textbf{29.82} & 29.22 & 30.06 & \textbf{31.23} \\
        \rowcolor{gray!10} ViT-B/16 & \textbf{20.79} & 16.08 & 16.08 & \textbf{16.98} & \textbf{19.77} & 15.93 & \textbf{22.47} & 16.98 & \textbf{18.27} & 17.43 & \textbf{18.66} & 16.98 \\
        Swin-T & \textbf{42.36} & 31.71 & 9.75 & \textbf{38.67} & \textbf{41.55} & 35.64 & \textbf{44.76} & 38.67 & \textbf{41.82} & 37.74 & \textbf{40.74} & 38.67 \\
        \bottomrule
    \end{tabular}%
    }
\end{table}


\subsubsection{Hierarchical per-class analysis}
\label{subsubsec:hierarchical}

Top-1 accuracy measures only whether the model identifies the correct leaf; it
does not capture whether mistakes respect the structure of the taxonomy.
To examine this, we evaluate per-class accuracy at the \emph{family} level (one above the leaves) of
the FGVC hierarchy, using the ResNet-50 models from Table~\ref{tab:besttobest}.
Leaf probabilities are propagated upward to the family nodes and the node with
the highest accumulated mass is selected; a prediction is correct if it matches
the family ancestor of the true class.

\begin{figure}
    \centering
    \includegraphics[width=1\linewidth]{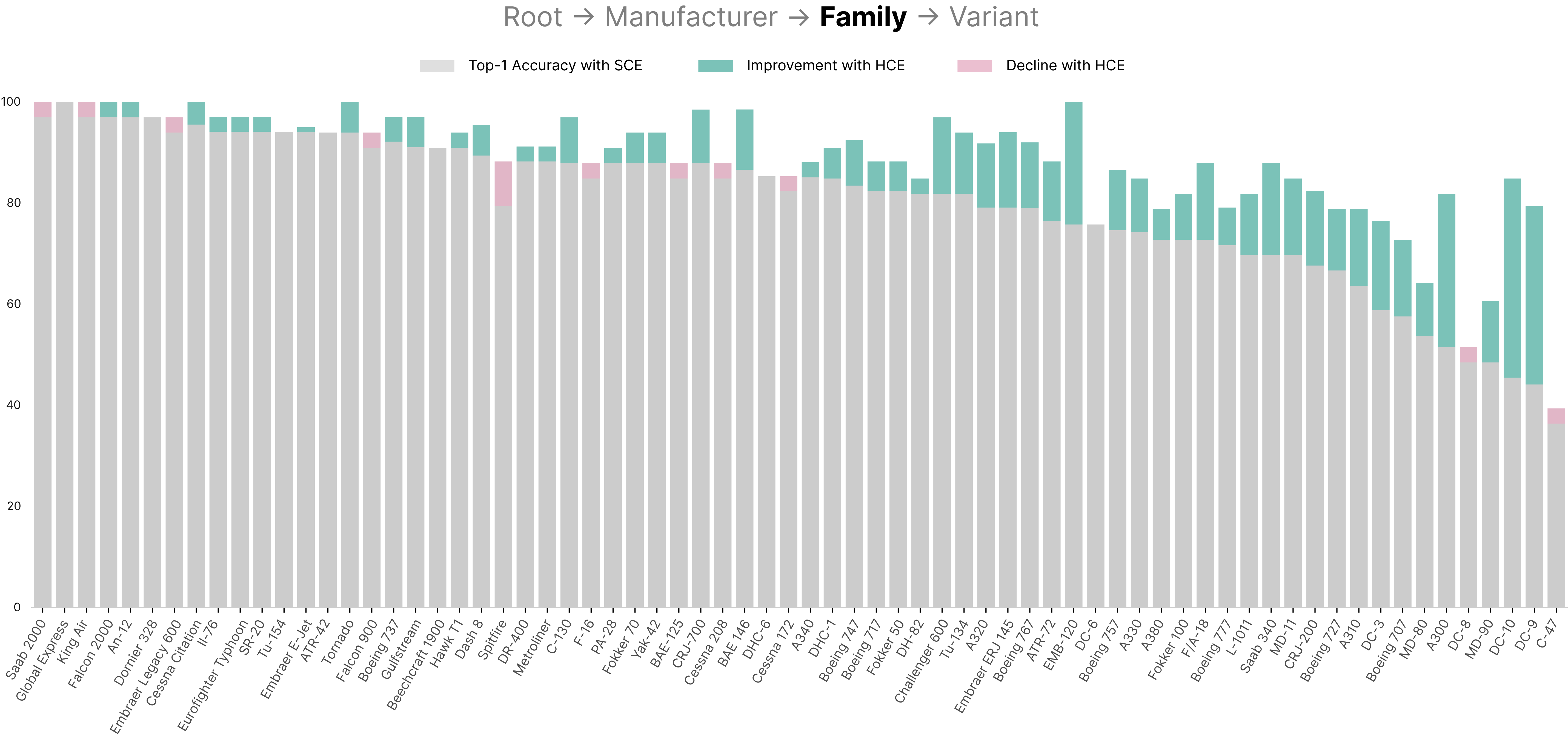}
    \caption{Per-class accuracy at the family level of the FGVC Aircraft
    hierarchy, comparing HACE and SCE on ResNet-50. Each bar shows the top-1
    accuracy achieved by SCE (grey), with the additional gain from HACE shown
    in teal (improvement) or pink (decline). Classes are sorted in descending
    order of SCE accuracy. HACE improves over SCE in 59 out of 70 families
    (mean gain 7.21\%). For both models, internal-node predictions are obtained
    by aggregating leaf probabilities upward through the reachability matrix;
    HACE produces these natively, while for SCE the aggregation is applied
    post-hoc at evaluation time.}
    \label{fig:family_accuracy}
\end{figure}

Figure~\ref{fig:family_accuracy} shows the results.
HACE matches or outperforms SCE in 59 out of 70 families, with a mean gain of
7.21\%. Results at an even coarser level of the hierarchy, the \emph{manufacturer} level, are reported in Appendix~\ref{app:hierarchical_fgvc}, where HACE matches or outperforms SCE in 28 out of 30 classes.

\subsubsection{Comparative analysis}
\label{subsec:ablation}


To situate HACE within the broader landscape of hierarchy-aware and smoothing-based losses, we compare eight methods on NABirds under identical training conditions.
On the HACE side, we evaluate a base variant (with no horizontal smoothing) and a variant that adds standard label smoothing before ancestral redistribution.
On the SCE side, we include standard cross-entropy, SCE with uniform label smoothing~\cite{szegedy2016rethinking}, and SCE using the LCA-based soft targets of Bertinetto
et al.~\cite{bertinetto2020making}.
Finally, HXE~\cite{bertinetto2020making} is evaluated at two values of parameter $\alpha$.

As shown in Table~\ref{tab:ablation}, HACE with label smoothing achieves the best
performance in 4 out of 6 architectures, with a mean improvement of 4.38\% over the next best method per architecture.
The consistent gap between base HACE and the SCE variants (with and without smoothing) indicates that the gains are not explained by label smoothing alone.
The improvement of HACE with label smoothing over base HACE further shows that horizontal and vertical smoothing are complementary: standard label smoothing provides additional gains on top of ancestral smoothing.

\begin{table}
    \centering
    \caption{Top-1 accuracy (\%) on NABirds for six methods under identical
    training conditions. HACE: prediction aggregation and ancestral label
    smoothing ($\varepsilon = 0$). HACE Smooth: same, with additional standard
    label smoothing ($\varepsilon = 0.1$) applied before the ancestral
    redistribution. SCE and SCE Smooth: standard cross-entropy without and with
    uniform label smoothing. Soft Label~\cite{bertinetto2020making}:
    LCA-based soft targets without prediction aggregation.
    HXE~\cite{bertinetto2020making}: hierarchically-weighted cross-entropy
    from the leaf softmax. The best result per architecture is \textbf{bolded}
    and the second best is \underline{underlined}.}
    \label{tab:ablation}
    \resizebox{\textwidth}{!}{%
    \begin{tabular}{l cc cccc cc}
        \toprule
        & \multicolumn{2}{c}{\textbf{HACE}}
        & \multicolumn{4}{c}{\textbf{SCE}}
        & \multicolumn{2}{c}{\textbf{HXE}} \\
        \cmidrule(lr){2-3} \cmidrule(lr){4-7} \cmidrule(lr){8-9}
        & \multicolumn{1}{c}{\small \textbf{No smooth.}} & \multicolumn{1}{c}{\small  \textbf{Label smooth.}}
        & \multicolumn{1}{c}{\small \textbf{No smooth.}} & \multicolumn{1}{c}{\small  \textbf{Label smooth.}}
        & \multicolumn{1}{c}{\small \textbf{Soft label $\beta=10$}} & \multicolumn{1}{c}{\small  \textbf{Soft label $\beta=30$}}
        & \multicolumn{1}{c}{\small $\boldsymbol{\alpha=0.2}$} & \multicolumn{1}{c}{\small  $\boldsymbol{\alpha=0.5}$} \\
        
        \midrule

        \rowcolor{gray!10} ResNet-18  & \underline{67.93} & \textbf{70.81} & 62.94 & 66.57 & 67.04 & 59.21 & 61.73 & 65.98 \\
        ResNet-34  & 69.44 & \textbf{73.02} & 71.01 & \underline{72.15} & 71.01 & 62.36 & 66.07 & 70.62 \\
        \rowcolor{gray!10} ResNet-50  & 67.90 & \textbf{73.15} & 59.07 & \underline{71.47} & 70.41 & 59.98 & 63.06 & 69.57 \\
        ConvNeXt-T   & \underline{50.74} & \textbf{59.49} & 30.35 & 40.33 & 0.94 & 1.00 & 0.59 & 0.47 \\
        \rowcolor{gray!10} ViT-B/16   & \textbf{29.16} & 17.53 & \underline{27.44} & 23.14 & 15.64 & 17.50 & 18.78 & 17.72 \\
        Swin-T     & 44.01 & 47.10 & 41.51 & \underline{47.27} & \textbf{48.00} & 42.00 & 43.06 & 44.29 \\
        \bottomrule
    \end{tabular}%
    }
\end{table}


\subsection{Linear probing}
\label{subsec:finetuning}
To complement our end-to-end training results, we evaluate HACE in a fine-tuning setting using 
DINOv2-Large as a frozen backbone. For each dataset, we pass all images through DINOv2-Large 
once and extract the unit-normalized classification token from the final hidden state, producing 
a fixed 1024-dimensional feature representation per image. On top of these frozen features, we 
train a single linear classifier for all combinations of dataset, dilution value, and loss function, 
using the same learning rate pairing scheme described in Appendix~\ref{app:lr}. 
Because the linear classifier cannot learn complex representations from fixed features, this setting 
provides a particularly clean test of the loss function's effect, disentangled from any interaction 
with feature learning.



Table~\ref{tab:finetuneresults} reports the best top-1 accuracy for each dataset across all dilution values and baselines. HACE outperforms all other methods in all 3 datasets, with an average improvement of 2.18\% over the next best (non-HACE) baseline per dataset.

\begin{table}[H]
    \centering
    \caption{Top-1 accuracy (\%) of all methods under linear probing on frozen
    DINOv2-Large features. Each cell reports the best result across all dilution
    values $d \in \{0.2, 0.5, 0.7\}$ and both learning rate pairings.
    The best result per dataset is \textbf{bolded} and the second best is \underline{underlined}.}
    \label{tab:finetuneresults}
    \resizebox{\textwidth}{!}{%
    \begin{tabular}{l cc cc cc cc cc}
        \toprule
        & \multicolumn{2}{c}{\textbf{\small No label smooth.}}
        & \multicolumn{2}{c}{\textbf{\small Label smooth.}}
        & \multicolumn{2}{c}{\textbf{\small Soft label $\beta=10$}}
        & \multicolumn{2}{c}{\textbf{\small Soft label $\beta=30$}}
        & \multicolumn{2}{c}{\textbf{HXE}} \\
        \cmidrule(lr){2-3} \cmidrule(lr){4-5} \cmidrule(lr){6-7}
        \cmidrule(lr){8-9} \cmidrule(lr){10-11}
        & \multicolumn{1}{c}{\textbf{HACE}} & \multicolumn{1}{c}{\textbf{SCE}}
        & \multicolumn{1}{c}{\textbf{HACE}} & \multicolumn{1}{c}{\textbf{SCE}}
        & \multicolumn{1}{c}{\textbf{HACE}} & \multicolumn{1}{c}{\textbf{SCE}}
        & \multicolumn{1}{c}{\textbf{HACE}} & \multicolumn{1}{c}{\textbf{SCE}}
        & \multicolumn{1}{c}{$\boldsymbol{\alpha=0.2}$} & \multicolumn{1}{c}{$\boldsymbol{\alpha=0.5}$} \\
        \midrule
        \rowcolor{gray!10} CIFAR-100 & \textbf{96.90} & 96.62 & \underline{96.81} & 96.68 & 93.07 & 92.90 & 93.10 & 92.95 & 92.72 & 92.52 \\
        FGVC      & \underline{58.97} & 53.91 & 58.70 & 53.40 & 58.90 & 53.83 & \textbf{58.99} & 54.10 & 49.81 & 42.90 \\
        \rowcolor{gray!10} NABirds   & \textbf{84.96} & 83.50 & \underline{84.78} & 83.02 & 84.70 & 83.00 & \textbf{84.96} & 83.50 & 79.09 & 64.86 \\
        \bottomrule
    \end{tabular}
    }
\end{table}

\subsubsection{Model-matched comparisons}

Table~\ref{tab:model-matchedfinetune} provides a per-dilution breakdown of HACE against SCE across all datasets. HACE outperforms SCE in 17 out of 18 dataset-dilution pairings, with an average improvement of 1.59\%, a maximum gain of 5.06\%, and a maximum decrease of 0.99\%. The smaller magnitude of improvements relative to end-to-end training is expected: all methods trained on CIFAR-100 already exceed 96\% accuracy, leaving little room for any method to demonstrate substantial gains. Across the more challenging FGVC and NABirds datasets, the advantage of HACE remains consistent.

\begin{table} [H]
    \centering
    \caption{Top-1 accuracy (\%) of HACE and SCE under linear probing on frozen DINOv2-Large features, broken down by dataset, dilution value $d$, and learning rate pairing (standard / adjusted). The better result per configuration is \textbf{bolded}.}
       \label{tab:model-matchedfinetune}
    \resizebox{\textwidth}{!}{%
    \begin{tabular}{l *{6}{cc}}
        \toprule
        & \multicolumn{2}{c}{$d = 0.2$}
        & \multicolumn{2}{c}{$d = 0.2$ adj}
        & \multicolumn{2}{c}{$d = 0.5$}
        & \multicolumn{2}{c}{$d = 0.5$ adj}
        & \multicolumn{2}{c}{$d = 0.7$}
        & \multicolumn{2}{c}{$d = 0.7$ adj} \\
        \cmidrule(lr){2-3} \cmidrule(lr){4-5} \cmidrule(lr){6-7}
        \cmidrule(lr){8-9} \cmidrule(lr){10-11} \cmidrule(lr){12-13}
        
            & HACE & SCE & HACE & SCE & HACE & SCE
            & HACE & SCE & HACE & SCE & HACE & SCE \\
        \midrule
        \rowcolor{gray!10} CIFAR-100 & \textbf{96.48} & 95.53 & \textbf{96.90} & 96.62 & \textbf{96.56} & 96.53 & \textbf{96.77} & 96.62 & \textbf{96.63} & 96.52 & \textbf{96.67} & 96.62 \\
        FGVC & \textbf{41.36} & 36.33 & \textbf{58.97} & 53.91 & \textbf{50.27} & 46.09 & \textbf{57.58} & 53.91 & \textbf{52.71} & 50.36 & \textbf{56.56} & 53.91 \\
        \rowcolor{gray!10} NABirds & \textbf{68.72} & 68.65 & \textbf{84.96} & 83.49 & \textbf{80.18} & 78.56 & \textbf{84.68} & 83.49 & 80.18 & \textbf{81.17} & \textbf{84.29} & 83.49 \\
        \bottomrule
    \end{tabular}%
    }
\end{table}

\section{Discussion}
\label{sec:discussion}
HACE consistently improves over standard cross-entropy across a diverse set of architectures, datasets, and training regimes, with gains that grow with the depth and irregularity of the class hierarchy. This pattern supports our central claim: when a known hierarchy is available, incorporating it directly into the loss is a simple and effective way to improve classification performance without any change to the model architecture or training pipeline.

Several limitations are worth noting. First, HACE requires a known class hierarchy, which may not always be available. Second, the dilution parameter $d$ introduces an additional hyperparameter; while we show consistent gains across a fixed set of values, dedicated tuning could yield further improvements. More broadly, a theoretical characterization of optimal ancestral smoothing functions, and the conditions under which they improve generalization, remains an open question.



\bibliographystyle{unsrtnat}
\bibliography{references}


\appendix

\section{Appendix}



\subsection{Extension to directed acyclic graphs}
\label{subsec:dag_extension}

The description of HACE in Section~\ref{sec:methods} assumes that the class hierarchy is a tree, so that every node has a unique parent
and the ancestral path from any leaf to the root is unique.
Both components of HACE extend naturally to directed acyclic graphs (DAGs), where a node may have multiple parents.

\textbf{Prediction aggregation} requires no modification: the reachability
matrix $\mathcal{R}$ is defined identically for DAGs, with $\mathcal{R}_{ij}
= 1$ whenever $j$ is reachable from $i$ by following directed edges.
The matrix--vector product $q^* = \mathcal{R}\,\hat{q}$ still accumulates
probability mass from all descendants, and hierarchical consistency
($q^*(i) \geq q^*(j)$ for any descendant $j$ of $i$) is preserved.

\textbf{Ancestral label smoothing} requires a small adjustment.
In a tree, the entire mass $(1 - d)^j$ reaching depth $j$ is passed to the
unique parent; in a DAG, a node at depth $j$ may have $k$ parents, and the
mass must be distributed among them.
The natural extension is to split the outgoing mass equally among all parents:
each parent of node $b_j$ receives a share of $(1-d)^j\,\tilde{p}(c)$
proportional to $1/k$, where $k$ is the number of parents of $b_j$.
Formally, let $\mathcal{P}(v)$ denote the set of parents of node $v$.
The contribution of leaf $c$ to an ancestor $v$ via a path
$b_0 = c, b_1, \ldots, b_m = v$ is:
\begin{equation}
  p^*(v) \mathrel{+}=
    d\,(1-d)^m \prod_{j=0}^{m-1} \frac{1}{|\mathcal{P}(b_{j+1})|}
    \;\tilde{p}(c),
\label{eq:dag_smoothing}
\end{equation}
where the product accounts for the cumulative splitting at each step along
the path.
Summing over all paths from $c$ to $v$ and all leaves $c$ with nonzero
$\tilde{p}(c)$ yields the full soft target $p^* \in \mathbb{R}^N$, which
remains a valid probability distribution: the total mass assigned to all nodes
is preserved by construction, since at each node the outgoing mass is
partitioned (not duplicated) across parents.

\subsection{Reachability Matrix}
\label{subsec: Reachability Matrix}

Both prediction aggregation and loss computation rely on the \emph{reachability
matrix} $\mathcal{R} \in \{0,1\}^{N \times N}$, where $N$ is the total number
of nodes in the class hierarchy (leaves and internal nodes combined).
Entry $\mathcal{R}_{ij}$ is defined as
\begin{equation}
  \mathcal{R}_{ij} =
  \begin{cases}
    1 & \text{if node } j \text{ is a descendant of node } i\text{, or } j = i, \\
    0 & \text{otherwise.}
  \end{cases}
\end{equation}
Equivalently, $\mathcal{R}$ is the reflexive transitive closure of the
parent-to-child adjacency matrix of the hierarchy: it encodes, for each node
$i$, the complete set of nodes whose raw softmax probability should be
accumulated into $\tilde{q}(i)$.

The aggregated prediction introduced in
Section~\ref{subsec:aggregated_prediction} can then be written as a single
matrix--vector product:
\begin{equation}
  q^* = \mathcal{R}\,\hat{q},
\end{equation}
where $\hat{q} \in \mathbb{R}^N$ is the raw softmax output over all $N$ nodes.
This makes aggregation a single differentiable operation with no explicit
tree traversal, and gradients flow back through $\mathcal{R}$ to the raw logits
during training.

\paragraph{Properties.}
The reachability relation encoded in $\mathcal{R}$ is a partial order: it is
reflexive (every node reaches itself), antisymmetric (the hierarchy is acyclic,
so if $i$ reaches $j$ and $j$ reaches $i$ then $i = j$), and transitive (if $i$
reaches $j$ and $j$ reaches $k$, then $i$ reaches $k$).
A direct consequence is that $q^*(i) \geq q^*(j)$ whenever $j$ is a descendant
of $i$, which is exactly the hierarchical consistency constraint: the probability
of any node is at least the probability of any of its subtrees.

\paragraph{Construction.}
For each dataset, $\mathcal{R}$ is built once before training by computing the
transitive closure of the parent-to-child adjacency matrix and setting the
diagonal to 1 to enforce reflexivity.
At the scales considered here, the matrix fits comfortably in memory as a dense Boolean tensor and can be moved to GPU for efficient batched matrix--vector products.

\subsection{Dataset Descriptions}
\label{app:datasets}

\paragraph{CIFAR-100~\cite{krizhevsky2009cifar}.}
CIFAR-100 contains 60{,}000 colour images at $32 \times 32$ pixels, split into
50{,}000 training and 10{,}000 test samples.
The 100 leaf classes are grouped into 20 superclasses, each covering exactly 5
leaves, giving a perfectly balanced two-level hierarchy (superclass $\to$ class).
The hierarchy is shallow and uniform: every leaf is at the same depth, and every
internal node has the same branching factor.
This makes CIFAR-100 a clean testbed for measuring whether HACE provides any
benefit when the hierarchy is simple and the images are low-resolution.

\paragraph{FGVC Aircraft~\cite{maji2013fgvc}.}
FGVC Aircraft contains 10{,}000 images of aircraft variants at $224\times224$
pixels.
The label hierarchy has three levels below the root: \emph{manufacturer}
(e.g.\ Boeing), \emph{family} (e.g.\ Boeing 7-series), and \emph{variant}
(e.g.\ Boeing 737-300), yielding a three-level tree.
The original taxonomy defines 214 nodes; we exclude 13 internal nodes that have
no associated leaf images (i.e.\ manufacturer or family nodes whose entire subtree
was removed from the dataset), yielding 201 nodes in total.
The hierarchy is moderately deep and unbalanced: different manufacturers have
different numbers of families, and families differ in the number of variants they
contain.
FGVC is the smallest dataset in our evaluation, which allows rapid experimentation
across the full grid of architectures and dilution values.

\paragraph{NABirds~\cite{vanhorn2015nabirds}.}
NABirds contains approximately 48{,}000 images of North American bird species at
$224\times224$ pixels, split into 23{,}929 training and 24{,}633 test images.
The 555 leaf classes (species variants) are organized into a biological taxonomy
that reflects standard ornithological classification.
Most species follow a three-level path below the root:
\emph{order} $\to$ \emph{species} $\to$ \emph{variant}.
However, the order \emph{Passeriformes} (perching birds) is exceptionally large,
accounting for over 60\% of all species in the dataset.
To reflect the finer-grained taxonomic structure within this order, the dataset
authors subdivide it by \emph{family} before reaching the species level, resulting
in a four-level path (order $\to$ family $\to$ species $\to$ variant).
This mixture of depths makes NABirds the most irregular hierarchy in our
evaluation: internal nodes have highly variable branching factors, leaf depth is
not uniform, and the taxonomy closely mirrors a real biological ontology rather
than a purpose-built classification scheme.




\subsection{Learning Rates}
\label{app:lr}

Ancestral label smoothing modifies the ground-truth target $p^*$ so that the
mass assigned to the true leaf is exactly $d$ (when $\varepsilon = 0$), rather
than $1$ as in a one-hot encoding.
Because the cross-entropy gradient with respect to the logit of the true class
is proportional to the target value at that class, the leaf-level gradient
contribution under HACE is exactly $d$ times the corresponding contribution
under SCE, for the same model output.
If both losses are trained at the same learning rate $\ell$, the leaf-level
weight updates differ by a factor of $d$, which is a confound we control for when comparing the two losses\footnote{Note that HACE also receives gradient signal from the internal nodes of the
hierarchy, which SCE does not; the learning rate adjustment discussed here
are motivated only by the leaf-level contribution, which is directly comparable between
the two losses.}.
We therefore consider two symmetric directions of adjustment reported in Table~\ref{tab:learning_rate}:
\begin{itemize}
    \item \textbf{HACE-anchored pairing (standard).} HACE is trained at the base rate $\ell$ recommended by the original authors. SCE is trained at $\ell \cdot d$, scaling its learning rate \emph{down} so that its leaf-level gradient contribution matches that of HACE.
    \item \textbf{SCE-anchored pairing (adj).} SCE is trained at the base rate $\ell$. HACE is trained at $\ell / d$, scaling its learning rate \emph{up} to match thelarger leaf-level gradient of SCE.
\end{itemize}

\begin{table}
  \centering
  \caption{Learning rate pairings used in experiments.
  The base learning rate $\ell$ is the value recommended by the original
  authors for each architecture (see Table~\ref{tab:models}).}
  \label{tab:learning_rate}
  \begin{tabular}{l l c}
    \toprule
    \textbf{Pairing} & \textbf{Loss}
      & \textbf{Learning Rate}\\
    \midrule
    \rowcolor{gray!10} HACE-anchored (standard) & HACE & $\ell$          \\
    HACE-anchored (standard) & SCE  & $\ell \cdot d$      \\
    \midrule
    \rowcolor{gray!10} SCE-anchored (adj)       & HACE & $\ell / d$  \\
    SCE-anchored (adj)       & SCE  & $\ell$          \\
    \bottomrule
  \end{tabular}
\end{table}

Each experiment is one triple (architecture, dataset, dilution value $d$).
HACE is trained under both pairings, giving $3 \times 2 = 6$ HACE models per
architecture--dataset pair.
For SCE, the HACE-anchored pairing uses a rate $\ell / d$ that depends on $d$,
so 3 distinct SCE models are needed; the SCE-anchored pairing always uses $\ell$
regardless of $d$, so a single SCE model covers all three dilution values.
This yields $3 + 1 = 4$ SCE models per architecture--dataset pair, for a total
of 10 models per pair.

\subsection{Ancestral Label Smoothing Matrix}
\label{subsec: Label Smoothing Matrix}

To avoid recomputing the soft target $p^*$ at every training step, we
precompute it for every leaf class and store the result as the \emph{ancestral
label smoothing matrix} $\mathcal{T} \in \mathbb{R}^{n \times N}$, where $n$ is
the number of leaf classes and $N$ is the total number of nodes.
Row $i$ of $\mathcal{T}$ contains the soft target $p^*$ for leaf class $i$,
constructed following the two-step procedure in
Section~\ref{subsec:label_smoothing} with the chosen values of $\varepsilon$
and $d$.

During training, the soft target for a batch of samples is retrieved by
indexing into $\mathcal{T}$ with the ground-truth label vector:
\[
  P^*_{\text{batch}} = \mathcal{T}[\mathbf{y}] \in \mathbb{R}^{B \times N},
\]
where $B$ is the batch size and $\mathbf{y} \in \{1,\ldots,n\}^B$ is the
vector of ground-truth leaf indices.
This reduces soft-target construction to a single gather operation per batch
with no runtime overhead beyond the one-time precomputation.



\subsection{Models}
\label{app:models}

We evaluate six architectures spanning two design families.
Table~\ref{tab:models} reports the hyperparameters used for each, along with the
number of training epochs per dataset.
All hyperparameters are set to the values recommended by the original authors
and held fixed across all loss functions and datasets; the only exception is the
learning rate, which is adjusted as described in Appendix~\ref{app:lr}.
This ensures that any observed differences in performance are attributable to the
choice of loss function alone.

\paragraph{Convolutional architectures.}
\textbf{ResNet-18, ResNet-34, ResNet-50}~\cite{he2016deep} are members of the
residual network family, trained with SGD and cosine annealing.
The three depths allow us to examine whether the benefit of HACE interacts with
model capacity within the same architectural family.
\textbf{ConvNeXt-Tiny}~\cite{liu2022convnet} is a modernized convolutional
design that incorporates design choices from transformers (depthwise convolutions,
layer normalization, inverted bottleneck blocks) while retaining a fully
convolutional structure; it is trained with AdamW.

\paragraph{Attention-based architectures.}
\textbf{ViT-B/16}~\cite{dosovitskiy2021an} is a pure Vision Transformer that
splits each image into $16 \times 16$ patches and processes them with global
self-attention; it uses dropout ($p = 0.1$) and is trained with AdamW.
\textbf{Swin-T}~\cite{liu2021swin} introduces a shifted window attention scheme
that restricts self-attention to local windows and progressively merges patches,
giving a hierarchical feature map similar to a convolutional backbone; it is
also trained with AdamW.

\paragraph{Linear probing backbone.}
For the linear probing experiments, we use
\textbf{DINOv2-Large}~\cite{oquab2024dinov2} as a frozen feature extractor.
DINOv2-Large is a ViT-L/14 ($\sim$300M parameters) trained with self-supervised
DINO objectives on a large curated dataset.
For each image, we extract the unit-normalized classification token from the
final hidden state, producing a fixed 1024-dimensional representation.
These features are computed once per dataset before any training begins.
Because the backbone is frozen throughout, this setting cleanly isolates the
effect of the loss function from any interaction with feature learning.

\begin{table}[!htbp]
  \centering
  \caption{Hyperparameters for the six end-to-end architectures.
  Epochs are reported as CIFAR-100 (C) / FGVC (F) / NABirds (N).}
  \label{tab:models}
  \resizebox{\textwidth}{!}{%
  \begin{tabular}{l c c c c c c c c c r}
    \toprule
    \textbf{Model} & \textbf{Type}
      & \textbf{Params} & \textbf{Batch} & \textbf{LR $\ell$}
      & \textbf{Weight Decay} & \textbf{Momentum} & \textbf{Dropout}
      & \textbf{Epochs (C\,/\,F\,/\,N)}
      & \textbf{LR Scheduler} & \textbf{Optimizer} \\
    \midrule
    \rowcolor{gray!10}
    ResNet-18   & Conv & $\sim$11.7M & 128 & $0.1$
      & $10^{-4}$ & $0.9$ & ---
      & 450\,/\,750\,/\,750 & CosineAnnealing & SGD \\
    ResNet-34   & Conv & $\sim$21.8M & 128 & $0.1$
      & $10^{-4}$ & $0.9$ & ---
      & 450\,/\,750\,/\,750 & CosineAnnealing & SGD \\
    \rowcolor{gray!10}
    ResNet-50   & Conv & $\sim$25.6M & 128 & $0.1$
      & $10^{-4}$ & $0.9$ & ---
      & 450\,/\,750\,/\,750 & CosineAnnealing & SGD \\
    ConvNeXt-T  & Conv & $\sim$28.6M & 128 & $1.25\times10^{-4}$
      & $0.05$ & $0.9$ & ---
      & 400\,/\,400\,/\,500 & CosineAnnealing & AdamW \\
    \rowcolor{gray!10}
    Swin-T      & Attn & $\sim$28.3M & 128 & $1.25\times10^{-4}$
      & $0.05$ & $0.9$ & ---
      & 500\,/\,500\,/\,500 & Lambda & AdamW \\
    ViT-B/16    & Attn & $\sim$86M &  64 & $3\times10^{-3}$
      & $0.3$ & $0.9$ & $0.1$
      & 300\,/\,300\,/\,300 & Lambda & AdamW \\
    \bottomrule
  \end{tabular}}
\end{table}

\subsection{Top-5 Accuracy}
\label{app:top5}

Tables~\ref{tab:top5_besttobest},~\ref{tab:results_nabirds_top5} and \ref{tab:results_top5_fine_tune} report top-5 accuracy for, respectively:
(i) end-to-end training across all architecture--dataset pairs (best HACE vs best SCE per pair), (ii) a controlled comparison of HACE, SCE, HXE, and soft-label variants on NABirds under identical training conditions, and (iii) linear probing on frozen DINOv2-Large features,
following the same format as the top-1 tables in the main paper. In end-to-end training, HACE outperforms SCE in 17 out of 18 architecture--dataset pairs, with a mean improvement of 4.46\%. In linear probing, HACE achieves the highest and second-highest top-5 accuracy
on all three datasets.

\begin{table}[!htbp]
  \centering
  \caption{Top-5 accuracy (\%) of HACE and SCE in end-to-end training, for each
  architecture--dataset pair. Each cell reports the best result across all
  dilution values $d \in \{0.2, 0.5, 0.7\}$ and both learning rate pairings.
  The better result per pair is \textbf{bolded}.}
  \label{tab:top5_besttobest}
  \begin{tabular}{l cc cc cc}
    \toprule
    & \multicolumn{2}{c}{\textbf{CIFAR-100}}
    & \multicolumn{2}{c}{\textbf{FGVC}}
    & \multicolumn{2}{c}{\textbf{NABirds}} \\
    \cmidrule(lr){2-3} \cmidrule(lr){4-5} \cmidrule(lr){6-7}
    \textbf{Architecture}
    & \multicolumn{1}{c}{HACE} & \multicolumn{1}{c}{SCE}
    & \multicolumn{1}{c}{HACE} & \multicolumn{1}{c}{SCE}
    & \multicolumn{1}{c}{HACE} & \multicolumn{1}{c}{SCE} \\
    \midrule
    \rowcolor{gray!10}
    ResNet-18 & \textbf{93.76} & 93.01 & \textbf{94.15} & 88.63 & \textbf{86.86} & 83.73 \\
    ResNet-34 & \textbf{94.87} & 93.77 & \textbf{93.97} & 93.64 & 87.67 & \textbf{88.08} \\
    \rowcolor{gray!10}
    ResNet-50 & \textbf{95.84} & 94.63 & \textbf{95.23} & 90.58 & \textbf{87.01} & 80.73 \\
    ConvNeXt-T  & \textbf{68.88} & 67.17 & \textbf{66.34} & 57.01 & \textbf{73.79} & 53.42 \\
    \rowcolor{gray!10}     ViT-B/16  & \textbf{85.29} & 85.25 & \textbf{45.21} & 33.99 & \textbf{53.34} & 51.23 \\
    Swin-T    & \textbf{83.06} & 80.04 & \textbf{73.06} & 66.31 & \textbf{69.25} & 66.16 \\
    \bottomrule
  \end{tabular}
\end{table}

\begin{table}[!htbp]
  \centering
  \caption{Top-5 accuracy (\%) on NABirds for all methods under identical
  training conditions (same as Table~\ref{tab:ablation}).
  The best result per architecture is \textbf{bolded} and the second best is
  \underline{underlined}.}
  \label{tab:results_nabirds_top5}
  \resizebox{\textwidth}{!}{%
  \begin{tabular}{l cc cccc cc}
    \toprule
    & \multicolumn{2}{c}{\textbf{HACE}}
    & \multicolumn{4}{c}{\textbf{SCE}}
    & \multicolumn{2}{c}{\textbf{HXE}} \\
    \cmidrule(lr){2-3} \cmidrule(lr){4-7} \cmidrule(lr){8-9}
    & \multicolumn{1}{c}{\small\textbf{No smooth.}}
    & \multicolumn{1}{c}{\small\textbf{Label smooth.}}
    & \multicolumn{1}{c}{\small\textbf{No smooth.}}
    & \multicolumn{1}{c}{\small\textbf{Label smooth.}}
    & \multicolumn{1}{c}{\small\textbf{Soft label $\beta=10$}}
    & \multicolumn{1}{c}{\small\textbf{Soft label $\beta=30$}}
    & \multicolumn{1}{c}{\small$\boldsymbol{\alpha=0.2}$}
    & \multicolumn{1}{c}{\small$\boldsymbol{\alpha=0.5}$} \\
    \midrule
    \rowcolor{gray!10}
    ResNet-18 & \underline{86.86} & \textbf{87.70} & 83.73 & 84.35 & 84.70 & 80.03 & 82.83 & 85.54 \\
    ResNet-34 & 87.67 & \textbf{89.38} & 88.08 & \underline{88.93} & 88.09 & 82.73 & 85.53 & 88.64 \\
    \rowcolor{gray!10}
    ResNet-50 & 87.01 & \textbf{89.65} & 80.73 & \underline{88.22} & 87.87 & 81.35 & 83.49 & 87.76 \\
    ConvNeXt-T  & \underline{73.79} & \textbf{79.04} & 53.42 & 62.78 & 4.44  & 4.54  & 3.00  & 2.23  \\
    \rowcolor{gray!10}    ViT-B/16  & \textbf{53.34} & 35.69 & \underline{51.23} & 44.52 & 33.05 & 36.05 & 38.10 & 36.80 \\
    Swin-T    & 69.25 & 70.01 & 66.16 & \underline{70.15} & \textbf{71.90} & 66.57 & 68.17 & 69.67 \\
    \bottomrule
  \end{tabular}}
\end{table}

\begin{table}[!htbp]
  \centering
  \caption{Top-5 accuracy (\%) of all methods under linear probing on frozen
  DINOv2-Large features. Each cell reports the best result across all dilution
  values $d \in \{0.2, 0.5, 0.7\}$ and both learning rate pairings.
  The best result per dataset is \textbf{bolded} and the second best is
  \underline{underlined}.}
  \label{tab:results_top5_fine_tune}
  \resizebox{\textwidth}{!}{%
  \begin{tabular}{l cc cc cc cc cc}
    \toprule
    & \multicolumn{2}{c}{\textbf{No label smooth.}}
    & \multicolumn{2}{c}{\textbf{Label smooth.}}
    & \multicolumn{2}{c}{\textbf{Soft label $\beta=10$}}
    & \multicolumn{2}{c}{\textbf{Soft label $\beta=30$}}
    & \multicolumn{2}{c}{\textbf{HXE}} \\
    \cmidrule(lr){2-3} \cmidrule(lr){4-5} \cmidrule(lr){6-7}
    \cmidrule(lr){8-9} \cmidrule(lr){10-11}
    & \multicolumn{1}{c}{\textbf{HACE}} & \multicolumn{1}{c}{\textbf{SCE}}
    & \multicolumn{1}{c}{\textbf{HACE}} & \multicolumn{1}{c}{\textbf{SCE}}
    & \multicolumn{1}{c}{\textbf{HACE}} & \multicolumn{1}{c}{\textbf{SCE}}
    & \multicolumn{1}{c}{\textbf{HACE}} & \multicolumn{1}{c}{\textbf{SCE}}
    & \multicolumn{1}{c}{$\boldsymbol{\alpha=0.2}$}
    & \multicolumn{1}{c}{$\boldsymbol{\alpha=0.5}$} \\
    \midrule
    \rowcolor{gray!10}
    CIFAR-100 & 99.03 & 98.94 & \textbf{99.05} & 98.97
              & 99.02 & 98.99 & \underline{99.04} & 98.96 & 98.96 & 98.92 \\
    FGVC      & \underline{86.02} & 81.43 & 84.73 & 81.10
              & \textbf{86.56} & 81.76 & \underline{86.02} & 81.47 & 78.07 & 73.11 \\
    \rowcolor{gray!10}
    NABirds   & \underline{97.60} & 97.21 & \underline{97.60} & 97.11
              & \textbf{97.62} & 97.23 & \underline{97.60} & 97.21 & 96.02 & 88.35 \\
    \bottomrule
  \end{tabular}}
\end{table}

\subsection{Manufacturer-level analysis on FGVC}
\label{app:hierarchical_fgvc}

For completeness, we report the same hierarchical per-class analysis at the
\emph{manufacturer} level of the FGVC Aircraft hierarchy.
The evaluation protocol mirrors that of Section~\ref{subsubsec:hierarchical}:
leaf probabilities are aggregated upward to manufacturer nodes and the node
with the highest accumulated mass is selected; a prediction is correct if it
matches the manufacturer ancestor of the true class.

\begin{figure}[h!]
    \centering
    \includegraphics[width=1\linewidth]{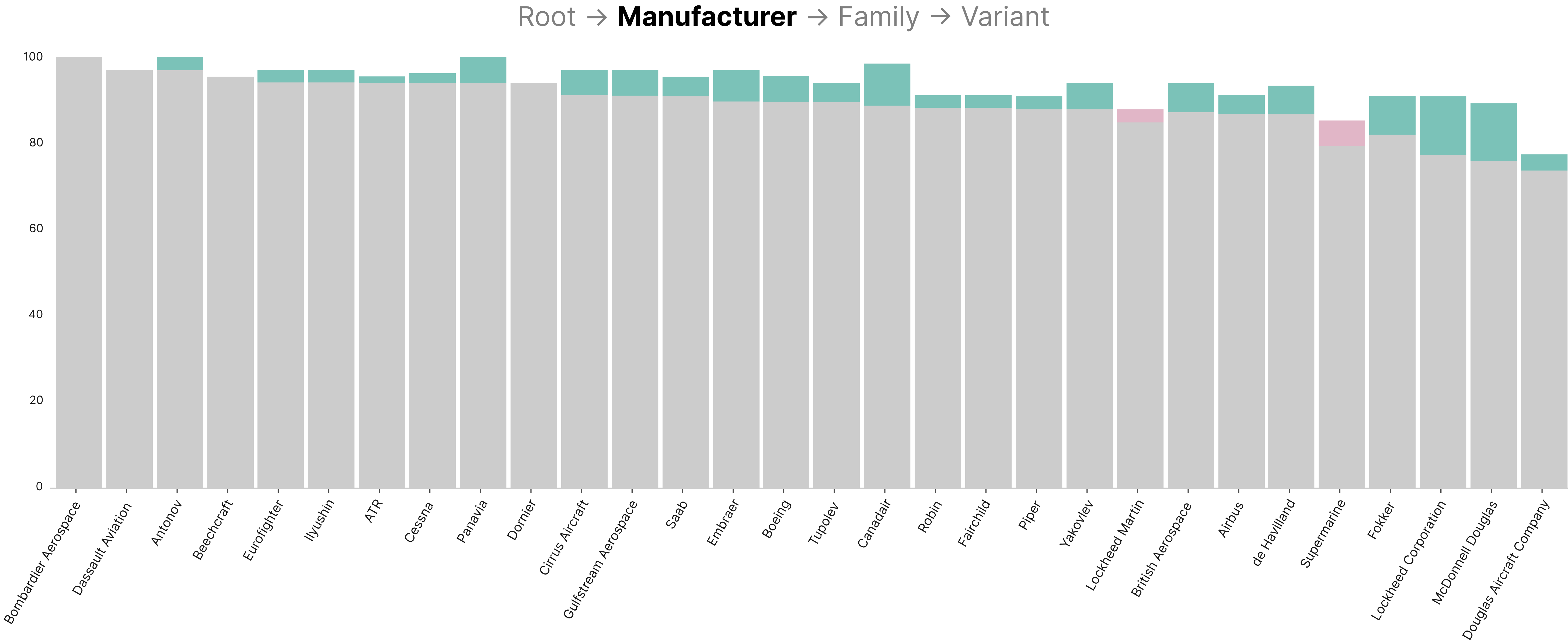}
    \caption{Per-class accuracy at the manufacturer level of the FGVC Aircraft
    hierarchy, comparing HACE and SCE on ResNet-50. Each bar shows the top-1
    accuracy achieved by SCE (grey), with the additional gain from HACE shown
    in teal (improvement) or pink (decline). Classes are sorted in descending
    order of SCE accuracy. HACE improves over SCE in 28 out of 30 manufacturers
    (mean gain 4.21\%).}
    \label{fig:manufacturer_accuracy}
\end{figure}

At this coarser level, HACE matches or outperforms SCE in 28 out of 30
manufacturers, with a mean gain of 4.21\%.
Figure~\ref{fig:manufacturer_accuracy} visualizes the per-class differences.

\newpage

\end{document}